\documentclass{bvm} 

\begin{document}

\selectlanguage{english} 

\title{Attention-based Multiple Instance Learning for Survival Prediction on Lung Cancer Tissue Microarrays}


\titlerunning{Attention-based MIL for Survival Prediction}

\newif\ifsubmission
\submissiontrue

\author{
	Jonas \lname{Ammeling} \inst{1}, 
	  Lars-Henning \lname{Schmidt} \inst{2},
	Jonathan \lname{Ganz} \inst{1}, 
        Tanja \lname{Niedermair} \inst{3},
        Christoph \lname {Brochhausen-Delius} \inst{3},
        Christian \lname{Schulz} \inst{4},
	Katharina \lname{Breininger} \inst{5},
	Marc \lname{Aubreville} \inst{1}
}


\authorrunning{Ammeling et al.}

\institute{
\inst{1} Technische Hochschule Ingolstadt, Ingolstadt, Germany\\
\inst{2} Medical Department IV, Pulmonary Medicine and Thoracic Oncology, Klinikum Ingolstadt, Ingolstadt, Germany\\
\inst{3} Institute of Pathology, University of Regensburg, Regensburg, Germany\\
\inst{4} Department of Internal Medicine II, University Hospital Regensburg, Regensburg, Germany \\
\inst{5} Department Artificial Intelligence in Biomedical Engineering, Friedrich-Alexander-Universität Erlangen-Nürnberg, Erlangen, Germany\\
}

\email{jonas.ammeling@thi.de}
\maketitle

\begin{abstract}
Attention-based multiple instance learning (AMIL) algorithms have proven to be successful in utilizing gigapixel whole-slide images (WSIs) for a variety of different computational pathology tasks such as outcome prediction and cancer subtyping problems. We extended an AMIL approach to the task of survival prediction by utilizing the classical Cox partial likelihood as a loss function, converting the AMIL model into a nonlinear proportional hazards model. We applied the model to tissue microarray (TMA) slides of 330 lung cancer patients. The results show that AMIL approaches can handle very small amounts of tissue from a TMA and reach similar C-index performance compared to established survival prediction methods trained with highly discriminative clinical factors such as age, cancer grade, and cancer stage.
\end{abstract}

\section{Introduction}

Despite advances in medicine over the past two decades, lung cancer remains the leading cause of cancer-related deaths worldwide \cite{Wang2021}. Survival prediction methods are used for predicting time-to-event outcomes such as cancer recurrence or death and play an important role in clinical decision making in oncology. Recent survival prediction methods used tissue samples from whole sections \cite{Chen2022, Silva2021} to make predictions about a patient's prognosis. However, with the trend toward minimally invasive biopsy techniques in the treatment of lung cancer, approaches that can process smaller amounts of tissue are needed \cite{Coley2015}. TMAs provide the opportunity to explore such methods because they typically contain only a very small amount of tissue for each patient from the original tumor sample \cite{Schmidt2009}. Bychkov et al. \cite{Bychkov2018} used a combination of convolutional and recurrent architectures to predict five-year survival as a binary classification problem from TMA cores of colon cancer patients. However, recurrent architectures are more complex than attention-based approaches, which have been shown to be successful for predicting patient prognosis on whole-slide images (WSI) \cite{Chen2022, Yao2020}. To investigate the applicability of attention-based methods to TMA slides, we extended an AMIL approach originally developed for weakly supervised classification to the survival prediction task similar to \cite{Chen2022} by converting the classification problem into a regression problem by optimizing the modified partial Cox likelihood \cite{Cox1972} as our loss function. We applied the model to TMA slides from patients with non-small cell lung cancer (NSCLC) and compared performance with established survival prediction methods trained on prognostically discriminative tabular data. The full code is available online\footnote{\label{3315-github}\url{https://github.com/DeepPathology/Cox_AMIL}}.


\section{Materials and Methods}


\subsection{Data}


Tissue samples, clinicopathologic features, and follow-up information were collected and examined from 379 NSCLC patients in the thoracic department of St. Georgs Klinikum in Ostercappeln, Germany. Clinical TNM staging (including clinical examination, CT scans, sonography, endoscopy, MRI, bone scan) was performed based on UICC/AJCC recommendations. The definite tumor staging was carried out post-surgically by pathological exploration. The histologic grade was determined according to WHO criteria on the original whole section tumor sample. The follow-up time was computed from the date of histological diagnosis to death or censored at the date of last contact. The TMAs were constructed by sampling with a 0.6 mm core needle from the most representative parts of the original tumor block. Three cores per patient were punched from the paraffin-embedded tumor block and assembled into TMA blocks. Multiple (at most 3x) four-micrometer-thick sections were cut from the TMA blocks, stained with hematoxylin and eosin (H\&E), and digitized with a whole-slide scanner (Pannoramic P1000, 3D HISTECH Ltd., Budapest, Hungary) at a resolution of $0.24$ $\mu$m/pixel. The TMA cores were linked to a unique patient identifier and extracted from the whole-slide TMA image. Due to missing values in the tabular data and loss of tissue sections from the TMA blocks, 49 cases were excluded from the analysis. Thus, data from 330 NSCLC patients was included for further analysis.

\subsection{Image Processing}

To compare the use of the patient-level label for different tissue amounts, all cores from the same patient were processed once individually and stitched together once horizontally to create a new patient-level TMA (Fig.~\ref{3315-arch}). All images were processed using the public CLAM \cite{Lu2021} repository for WSI analysis. After tissue segmentation, non-overlapping patches of size $256 \times 256$ were extracted at full resolution. Then, a ResNet50 model, pre-trained on ImageNet, was used to convert each patch into a $1024$-dimensional feature vector by spatial average pooling after the third residual block.

\subsection{Attention-based Multiple Instance Learning}

\begin{figure}[b]
	\includegraphics[width=0.9\textwidth]{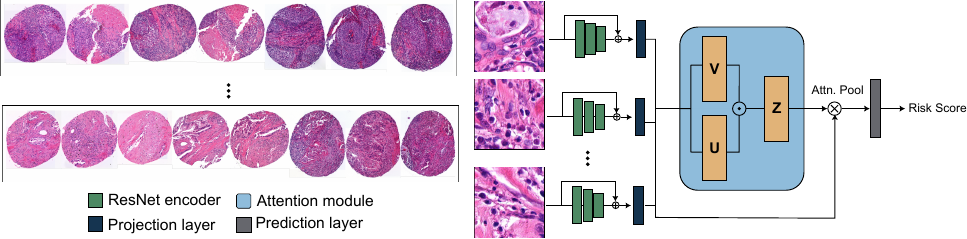}
	\caption{Data stitching and overall architecture. Left: tissue-cores of the same patient stitched together horizontally. Right: overview of the attention-based multiple instance learning algorithm for survival prediction.}
	\label{3315-arch} 
\end{figure}

 In the context of multiple instance learning, each image is viewed as a collection of $M$ patches or instances, known as a bag, with a corresponding bag-level label associated with it. After image processing, each bag is represented by the patch-level embeddings $\mathbf{H} \in \mathbb{R}^{M \times C}$, where $C$ is the feature dimension from the ResNet50 model. The model consists of three components as shown in Figure \ref{3315-arch}: the projection layer $f_\text{proj}$, the attention module $f_\text{attn}$, and the prediction layer $f_\text{pred}$. The projection layer $f_\text{proj}$ is a fully-connected layer with weights $\mathbf{W}_{\text{proj}} \in \mathbb{R}^{512 \times C}$ (all bias terms are implied for notational convenience) that maps the patch-level embedding into a more compact, dataset-specific 512-dimensional feature space. The attention module $f_\text{attn}$ learns to assign a score to each patch based on its contribution to the patient's predicted prognosis. In particular, $f_\text{attn}$ consists of three fully connected layers with weights $\mathbf{U} \in \mathbb{R}^{256 \times 512}$, $\mathbf{V} \in \mathbb{R}^{256 \times 512}$, and $\mathbf{Z} \in \mathbb{R}^{1 \times 256}$. After projection, the attention score $a_m$ for the $m$-th patch embedding $\mathbf{h}_{m} \in \mathbb{R}^{512}$, is computed by \cite{Lu2021}:

\begin{equation}
    a_m = \frac{\exp\bigl\{\mathbf{Z}\left(\tanh\left(\mathbf{V}\mathbf{h}_{m}^{\text{T}}\right) \odot \text{sigm}\left(\mathbf{U}\mathbf{h}_{m}^{\text{T}}\right)\right)\bigr\}}{\sum_{m=1}^{M} \exp\bigl\{\mathbf{Z}\left(\tanh\left(\mathbf{V}\mathbf{h}_{m}^{\text{T}}\right) \odot \text{sigm}\left(\mathbf{U}\mathbf{h}_{m}^{\text{T}}\right)\right)\bigr\}}
\end{equation}
 
The computed attention scores for each patch are then used as weight coefficients to aggregate the patch-level embeddings into the bag representation $\mathbf{h}_\text{bag} \in \mathbb{R}^{512}$ via attention-pooling \cite{Lu2021} by:

\begin{equation}
    \mathbf{h}_\text{bag} = \sum_{m=1}^{M}a_m\mathbf{h}_m
\end{equation}

The final prediction layer $f_\text{pred}$ is a single fully-connected layer with weights $\mathbf{W}_\text{pred}\in\mathbb{R}^{1 \times 512}$ with a single output node and linear activation. Let $\theta$ represent all weights of the network, then the model can be described as a function $h_\theta: \mathbb{R}^{M \times C} \mapsto \mathbb{R}$, where the output is a patient's hazard log risk score, described in more detail in the next section.

\subsection{Loss Function}

The right-censored patient-level survival data for the $i$-th patient consist of the triple $(t_i, \delta_i, x_i)$ and represent the observed time, binary censoring status ($\delta = 0$, death not observed), and image data, respectively. Censoring is assumed to be non-informative, such that for a given $x_i$, survival time and censoring are independent. Let $t_1 < t_2 < t_D$ be the ordered event times. The risk set $\mathcal{R}(t_i)$ is defined as the set of all individuals still in the study at a time immediately preceding $t_i$. The loss function is derived from the common Cox proportional hazards model where the hazard function of the form $\lambda(t\mid x) = \lambda_0(t)\cdot e^{h(x)}$ is composed of the hazard baseline function $\lambda_0(t)$, and a risk score $r(x) = e^{h(x)}$. Our proposed model estimates a patient's log-risk score $\hat{h}_\theta(x)$ parameterized by the weights of the network $\theta$ such that the modified Cox partial likelihood becomes \cite{Cox1972}:

\begin{equation}
    L(\theta)=\prod_{i:\delta=1} \frac{\hat{r}_\theta(x_i)}{\sum_{j \in \mathcal{R}(t_i)} \hat{r}_\theta(x_j)} = \prod_{i:\delta=1} \frac{\exp\left(\hat{h}_\theta(x_i)\right)}{\sum_{j \in \mathcal{R}(t_i)} \exp\left(\hat{h}_\theta(x_j)\right)}
\end{equation}

The loss function which is minimized by the network is then obtained from the average of the negative partial log likelihood with regularization as shown below:

\begin{equation}
    l(\theta) = - \frac{1}{N} \sum_i \delta_i \left(\hat{h}_\theta(x_i) - \log \sum_{j \in \mathcal{R}(t_i)} \exp\left(\hat{h}_\theta(x_j)\right)\right) + \lambda \cdot \| \theta \|_1, 
\end{equation}

where $N$ is the number of patients with an observable event and $\lambda$ is the $l_1$ regularization parameter. Intuitively, the loss function penalizes discordance between the scores of higher-risk and lower-risk patients. Similar loss functions have been used previously by other authors \cite{Katzman2018, Yao2020}. Details about the training and hyperparameter settings are online\textsuperscript{\ref{3315-github}}.



\subsection{Baselines}

The performance of the proposed model was compared against three established survival analysis methods based on tabular data. The tabular data used to train these baseline methods consist of patient characteristics (age, sex, and smoking status) and clinical characteristics (cancer stage and grade). It should be noted that the latter require elaborate determination and are commonly accepted to be highly prognostic. The first baseline method was a classical linear Cox proportional hazards (CPH) model \cite{Cox1972}. The second was a random survival forest (RSF) \cite{Ishwaran2008}, a non-proportional, flexible and robust alternative to the classical CPH model. The third baseline method was DeepSurv \cite{Katzman2018}, a modern nonlinear, deep learning-based CPH model. Moreover, the proposed model was compared with a MIL method using classical max-pooling, and with the performance of an AMIL model trained on each patient core individually, using the respective patient-level label.

\subsection{Evaluation}

To evaluate and compare the predictive performance on the survival data, we performed a 10-fold cross-validation and measured Harrell's concordance index (C-index) \cite{Harrell1982}. The C-index indicates how well a model predicts the ranking of patients' death times, where large values of $\hat{h}_\theta(x_i)$ should be associated with small values of $t_i$ and vice versa. A value of $C = 0.5$ corresponds to the average C-index of a random model, whereas $C = 1$ corresponds to a perfect association. Patient stratification was assessed by assigning patients to either a high-risk or a low-risk group based on the median of the predicted risk score. Kaplan-Meier curves were constructed by pooling the risk predictions of the test folds and plotting them against their survival time. Logrank tests were performed to check for statistically significant differences (P-value $< 0.05$) between the two survival distributions. 




\section{Results}

\begin{figure}[b]
	\includegraphics[width=0.88\textwidth]{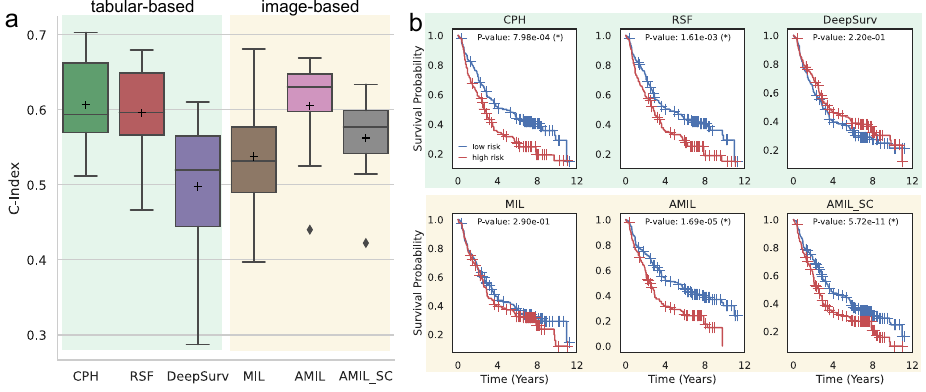}
	\caption{a) Cross-validated C-index performance. + indicates mean value. b) Kaplan-Meier curves for patient stratification results on the pooled validation splits.}
	\label{3315-results} 
\end{figure}

The cross-validated C-index values and the Kaplan-Meier curves for the patient stratification results are shown in Figure \ref{3315-results}. The classical CPH model achieved the overall best performance with an average C-index of $0.61\pm0.07$ (P-value: $7.98\times10^{-4}$) and a median C-index of $0.59$. Closely similar, the RSF model achieved an average C-index of $0.60\pm0.06$ (P-value: $1.61\times10^{-3}$) and a median C-index of $0.60$. The DeepSurv method performed the worst among all methods with an average C-index of $0.50\pm0.1$ (P-value: $2.20\times10^{-1}$) and a median C-index of $0.53$. Among the image-based methods, the classical MIL model with max-pooling performs the worst with an average C-index of $0.54\pm0.08$ (P-value: $2.90\times10^{-1}$) and a median C-index of $0.53$. The AMIL model performed the best among the image-based methods with an average C-index of $0.61\pm0.07$ (P-value: $1.69\times10^{-5}$) and a median C-index of $0.63$. The AMIL model trained on individual cores performed worse than the AMIL model trained using the patient label for all cores together, with an average C-index of $0.56\pm0.06$ (P-value: $5.72\times10^{-11}$) and a median C-index of $0.58$. 

\section{Discussion}

The results show that the proposed model performed similarly well to established survival prediction methods based on tabular data, such as CPH and RSF, by extracting features directly from the small amount of tissue provided by the TMA. Attention pooling proved to be very effective in this regard for aggregating these features to estimate the patient's risk score. However, the experiments also showed that the prognostically relevant information is not necessarily evenly distributed across a patient's tissue cores when selected from the original sample, as indicated by lower average C-index when trained on individual cores. Thus, performance is highly dependent on tissue core selection, and it is recommended to use all available tissue cores. Nevertheless, the results show that attention-based MIL approaches are applicable to TMA slides where only a small amount of tissue per patient is available to perform reliable patient stratification, as shown by the statistically significant logrank test results. Therefore, in the context of the trend toward minimally invasive biopsy procedures in lung cancer treatment, these methods could become a part of reliable decision support systems in the future.


\ifsubmission
    \begin{acknowledgement}
    	J. A. acknowledges funding by the Bavarian Institute for Digital Transformation (Project ReGInA). 
    \end{acknowledgement}
\else
\fi

\printbibliography

@article{Coley2015,
   author = {Shana M. Coley and John P. Crapanzano and Anjali Saqi},
   doi = {10.1002/CNCY.21527},
   issn = {1934-6638},
   issue = {5},
   journal = {Cancer Cytopathol},
   month = {5},
   pages = {318-326},
   pmid = {25711901},
   publisher = {John Wiley & Sons, Ltd},
   title = {FNA, core biopsy, or both for the diagnosis of lung carcinoma: Obtaining sufficient tissue for a specific diagnosis and molecular testing},
   volume = {123},
   url = {https://onlinelibrary.wiley.com/doi/full/10.1002/cncy.21527 https://onlinelibrary.wiley.com/doi/abs/10.1002/cncy.21527 https://acsjournals.onlinelibrary.wiley.com/doi/10.1002/cncy.21527},
   year = {2015},
}

@article{Bychkov2018,
   author = {Dmitrii Bychkov and Nina Linder and Riku Turkki and Stig Nordling and Panu E. Kovanen and Clare Verrill and Margarita Walliander and Mikael Lundin and Caj Haglund and Johan Lundin},
   doi = {10.1038/s41598-018-21758-3},
   issn = {2045-2322},
   issue = {1},
   journal = {Sci Rep},
   month = {2},
   pages = {1-11},
   pmid = {29467373},
   publisher = {Nature Publishing Group},
   title = {Deep learning based tissue analysis predicts outcome in colorectal cancer},
   volume = {8},
   url = {https://www.nature.com/articles/s41598-018-21758-3},
   year = {2018},
}

@article{Schmidt2009,
   author = {Lars Henning Schmidt and Stefan Biesterfeld and Andreas Kümmel and Andreas Faldum and Martin Sebastian and Christian Taube and Roland Buhll and Rainer Wiewrodt},
   issn = {0250-7005},
   issue = {1},
   journal = {Anticancer res},
   month = {1},
   pages = {201-9},
   pmid = {19331151},
   title = {Tissue microarrays are reliable tools for the clinicopathological characterization of lung cancer tissue.},
   volume = {29},
   year = {2009},
}

@article{Wang2021,
   author = {Meina Wang and Roy S. Herbst and Chris Boshoff},
   doi = {10.1038/s41591-021-01450-2},
   issn = {1078-8956},
   issue = {8},
   journal = {Nat Med},
   month = {8},
   pages = {1345-1356},
   title = {Toward personalized treatment approaches for non-small-cell lung cancer},
   volume = {27},
   year = {2021},
}

@article{Harrell1982,
   author = {Frank E. Harrell and Robert M. Califf and David B. Pryor and Kerry L. Lee and Robert A. Rosati},
   doi = {10.1001/JAMA.1982.03320430047030},
   issn = {0098-7484},
   issue = {18},
   journal = {JAMA},
   month = {5},
   pages = {2543-2546},
   pmid = {7069920},
   publisher = {American Medical Association},
   title = {Evaluating the Yield of Medical Tests},
   volume = {247},
   url = {https://jamanetwork.com/journals/jama/fullarticle/372568},
   year = {1982},
}

@article{Ishwaran2008,
    author = {Hemant Ishwaran and Udaya B. Kogalur and Eugene H. Blackstone and Michael S. Lauer},
    title = {{Random survival forests}},
    volume = {2},
    journal = {Ann Appl Stat},
    number = {3},
    publisher = {Institute of Mathematical Statistics},
    pages = {841 -- 860},
    year = {2008},
    doi = {10.1214/08-AOAS169},
    URL = {https://doi.org/10.1214/08-AOAS169}
}

@article{Cox1972,
   author = {D. R. Cox},
   doi = {10.1111/J.2517-6161.1972.TB00899.X},
   issn = {2517-6161},
   issue = {2},
   journal = {J R Stat Soc Series B Stat Methodol},
   month = {1},
   pages = {187-202},
   publisher = {John Wiley & Sons, Ltd},
   title = {Regression Models and Life-Tables},
   volume = {34},
   url = {https://onlinelibrary.wiley.com/doi/full/10.1111/j.2517-6161.1972.tb00899.x https://onlinelibrary.wiley.com/doi/abs/10.1111/j.2517-6161.1972.tb00899.x https://rss.onlinelibrary.wiley.com/doi/10.1111/j.2517-6161.1972.tb00899.x},
   year = {1972},
}

@article{Katzman2018,
   author = {Jared L Katzman and Uri Shaham and Alexander Cloninger and Jonathan Bates and Tingting Jiang and Yuval Kluger},
   doi = {10.1186/s12874-018-0482-1},
   issn = {1471-2288},
   issue = {1},
   journal = {BMC Med Res Methodol},
   pages = {24},
   title = {DeepSurv: personalized treatment recommender system using a Cox proportional hazards deep neural network},
   volume = {18},
   url = {https://doi.org/10.1186/s12874-018-0482-1},
   year = {2018},
}

@article{Chen2022,
   author = {Richard J. Chen and Ming Y. Lu and Drew F.K. Williamson and Tiffany Y. Chen and Jana Lipkova and Zahra Noor and Muhammad Shaban and Maha Shady and Mane Williams and Bumjin Joo and Faisal Mahmood},
   doi = {10.1016/j.ccell.2022.07.004},
   issn = {15356108},
   issue = {8},
   journal = {Cancer Cell},
   month = {8},
   pages = {865-878.e6},
   title = {Pan-cancer integrative histology-genomic analysis via multimodal deep learning},
   volume = {40},
   year = {2022},
}

@article{Yao2020,
   author = {Jiawen Yao and Xinliang Zhu and Jitendra Jonnagaddala and Nicholas Hawkins and Junzhou Huang},
   doi = {https://doi.org/10.1016/j.media.2020.101789},
   issn = {1361-8415},
   journal = {Med Image Anal},
   pages = {101789},
   title = {Whole slide images based cancer survival prediction using attention guided deep multiple instance learning networks},
   volume = {65},
   url = {https://www.sciencedirect.com/science/article/pii/S1361841520301535},
   year = {2020},
}

@article{Silva2021,
   author = {Luís A Vale-Silva and Karl Rohr},
   doi = {10.1038/s41598-021-92799-4},
   issn = {2045-2322},
   issue = {1},
   journal = {Sci Rep},
   pages = {13505},
   title = {Long-term cancer survival prediction using multimodal deep learning},
   volume = {11},
   url = {https://doi.org/10.1038/s41598-021-92799-4},
   year = {2021},
}

@article{Lu2021,
   author = {Ming Y Lu and Drew F K Williamson and Tiffany Y Chen and Richard J Chen and Matteo Barbieri and Faisal Mahmood},
   doi = {10.1038/s41551-020-00682-w},
   issn = {2157-846X},
   issue = {6},
   journal = {Nat Biomed Eng},
   pages = {555-570},
   title = {Data-efficient and weakly supervised computational pathology on whole-slide images},
   volume = {5},
   url = {https://doi.org/10.1038/s41551-020-00682-w},
   year = {2021},
}

\end{document}